\documentclass[conference]{IEEEtran}
\IEEEoverridecommandlockouts
\usepackage{cite}
\usepackage{amsmath,amssymb,amsthm,amsfonts}
\usepackage{textcomp}
\usepackage{fontenc}
\usepackage{times}
\usepackage{soul}
\usepackage{url}
\usepackage{hyperref}
\usepackage[utf8]{inputenc}
\usepackage{graphicx}
\usepackage{booktabs}
\usepackage{multirow}
\usepackage[ruled,vlined]{algorithm2e}
\usepackage{xcolor}
\usepackage{dblfloatfix}    % To enable figures at the bottom of page

\newcommand{\para}[1]{\vspace{0.2cm}\paragraph*{\textbf{#1}}}

\def\BibTeX{{\rm B\kern-.05em{\sc i\kern-.025em b}\kern-.08em
    T\kern-.1667em\lower.7ex\hbox{E}\kern-.125emX}}
\begin{document}

\title{Progressive Transmission and Inference of \linebreak Deep Learning Models}

\author{\IEEEauthorblockN{Youngsoo Lee}
\IEEEauthorblockA{\textit{School of Computing, KAIST} \\
Republic of Korea \\
youngsoo.lee@kaist.ac.kr}
\and

\IEEEauthorblockN{Sangdoo Yun}
\IEEEauthorblockA{\textit{NAVER AI Lab} \\
Republic of Korea \\
sangdoo.yun@navercorp.com}
\and

\IEEEauthorblockN{Yeonghun Kim}
\IEEEauthorblockA{\textit{School of Computing, KAIST} \\
Republic of Korea \\
neutrinoant@kaist.ac.kr}
\and

\IEEEauthorblockN{Sunghee Choi}
\IEEEauthorblockA{\textit{School of Computing, KAIST} \\
Republic of Korea \\
sunghee@kaist.edu}
}

\DeclareRobustCommand{\IEEEauthorrefmark}[1]{\smash{\textsuperscript{\footnotesize #1}}}

% \author{
%     \IEEEauthorblockN{Youngsoo Lee\IEEEauthorrefmark{1}, Sangdoo Yun\IEEEauthorrefmark{2}, Yeonghun Kim\IEEEauthorrefmark{1}, Sunghee Choi\IEEEauthorrefmark{1}}
%     \IEEEauthorblockA{\IEEEauthorrefmark{1}School of Computing, KAIST}
%     \IEEEauthorblockA{\IEEEauthorrefmark{2}NAVER Corp.}
%     \{youngsoo.lee, neutrinoant\}@kaist.ac.kr, sangdoo.yun@navercorp.com, sunghee@kaist.edu
% }

\maketitle

\begin{abstract}
Modern image files are usually progressively transmitted and provide a preview before downloading the entire image for improved user experience to cope with a slow network connection. 
In this paper, with a similar goal, we propose a progressive transmission framework for deep learning models, especially to deal with the scenario where pre-trained deep learning models are transmitted from servers and executed at user devices (e.g., web browser or mobile).
Our progressive transmission allows inferring approximate models in the middle of file delivery, and quickly provide an acceptable intermediate outputs.
On the server-side, a deep learning model is divided and progressively transmitted to the user devices.
Then, the divided pieces are progressively concatenated to construct approximate models on user devices.
Experiments show that our method is computationally efficient without increasing the model size and total transmission time while preserving the model accuracy. 
We further demonstrate that our method can improve the user experience by providing the approximate models especially in a slow connection.
\end{abstract}

\begin{IEEEkeywords}
deep learning model transmission, deep learning model deployment, deep learning application, progressive transmission, user experience
\end{IEEEkeywords}

\section{Introduction}

\begin{figure}[t]
    \centering
    \includegraphics[width=.45\textwidth]{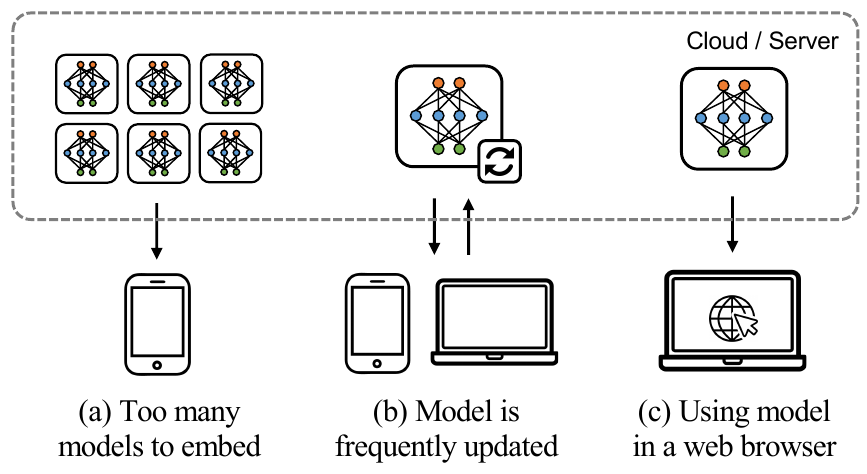}
    \caption{Scenarios where the deep learning models are transmitted over network.}
    \label{fig:scenarios}
\end{figure}

Recently, deep learning models have spread out to user edge devices beyond powerful enterprise servers \cite{wang2018towardsmobile}.
Although edge devices have relatively lower computational power than the servers, reasons such as data privacy, communication latency, or server-side computational burdens accelerate their adoption  \cite{TensorFlowJS,hidaka2017webdnn,zhang2019survey}.
One simple way to use deep learning models on edge devices is to embed the model in the application on the deployment stage.
However, it has not been considered as an issue when the model is transmitted over a network connection.
For example, in a complex application that uses multiple models, it might be impossible to embed all the models due to the limited storage resources (Fig. \ref{fig:scenarios}a).
In some cases, deep learning models might be transmitted again after the deployment since they are frequently updated in the server, or the user's situation-specific models might have to be provided each time (Fig. \ref{fig:scenarios}b).
On a web application, the deep models are executed on a web browser, which does not support pre-embedding of the models (Fig. \ref{fig:scenarios}c).
In these cases, user should wait for a long time until finishing the transmission.
And if the network connection is slow, the deep model transmission time would harm the user experience.
We further expect the problem might occur not only for a slow network connection but also for relatively fast network connection since the deep model size becomes larger for higher accuracy \cite{chen2018transmission}.

\begin{figure}[t]
    \centering
    \includegraphics[width=.48\textwidth]{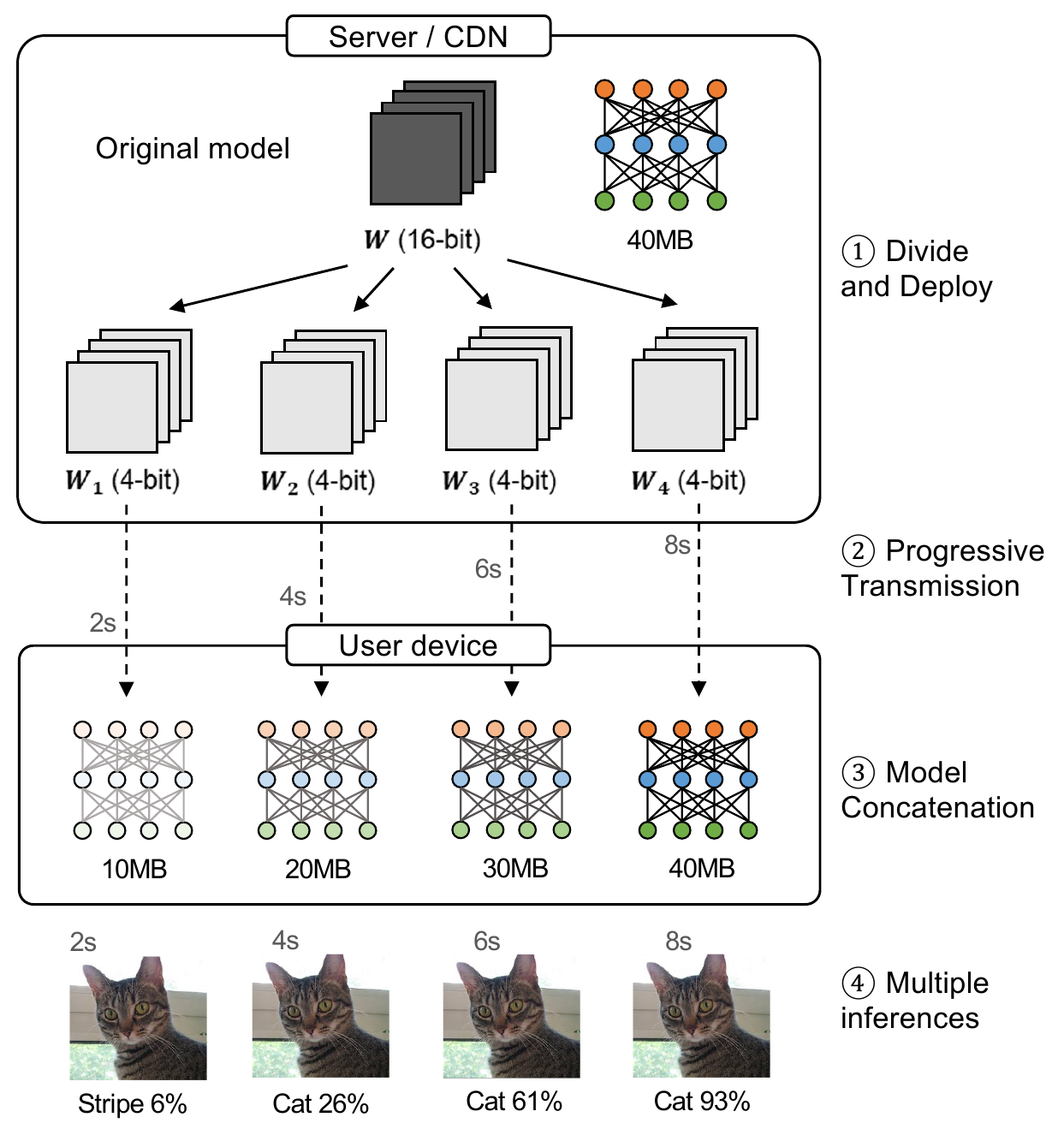}
    \caption{Flow illustration of progressive transmission and inference of deep learning models.}
    \label{fig:overview}
\end{figure}

In this paper, we introduce a compromise solution to the problem of long transmission time of deep learning models.
Instead of reducing the long transmission time itself, we propose to provide approximate inference results in the middle of the transmission.
A similar approach has been taken for image transmission, for example, the JPEG format supports progressive mode in which a reasonable preview is available after only a portion of the data is transmitted \cite{Sloan1979ProgressiveImage,Tzou1987ProgressiveImage}.
Inspired by the classic yet effective image transmission approach, we argue that progressive transmission of deep models could improve the response time and the user experience.
% Fig. \ref{fig:detection-example} shows an example of progressive transmission for the object detection model, in which the approximate detection results are provided in the middle of the model transmission.

Fig. \ref{fig:overview} illustrates the tasks and the flow for progressive transmission and inference.
As shown in the figure, the original model is divided before deployment, and concatenation and inference are conducted multiple times on the user device afterward.
To enable progressive transmission, an interface that supports flexible division and concatenation should be designed without increasing the overall model size.
In this paper, we present a new framework that includes division and concatenation schemes for progressive transmission and inference of deep learning models along with the compression algorithm.
Next, we demonstrate that our framework does not increase the model size and total execution time even if it provides approximate inference results.
We further evaluate whether our method improves user experience in a real-world web application by conducting a user study.
The source code and demo are available at {\url{https://github.com/Prev/progressivenet}}.

\section{Related Work}
\label{section:related_works}

\subsection{Progressive Transmission of Images}

The goal of classic progressive transmission of images, introduced by Sloan and Tanimoto \cite{Sloan1979ProgressiveImage} is to display a reasonable preview with minimum bits.
To display a reasonable approximate image, limiting the pixel values (i.e., quantizing the pixel value range from 256 to 16), or reducing the image resolution has been used \cite{Tzou1987ProgressiveImage}.
The algorithm used in image transmission cannot be used in the transmission of deep learning models since we transmit a completely different file types: a visual image and a bunch of matrices.
However, there is a common goal in that human perception is both considered when showing intermediate results.
In image transmission, the most plausible intermediate image must be shown.
In deep learning model transmission, providing an intermediate model with the highest possible accuracy is crucial.

\subsection{Deep Model Compression}

One way to improve user experience when transmitting a deep model over network is to compress the deep model. 
Many techniques have been proposed such as network pruning \cite{guo2016pruning,han2016deepcomp}, weight quantization \cite{Gong2014Quantization,Jacob2018QuantizationForIAI}, matrix factorization \cite{yu2017decomposition}, and knowledge distillation \cite{hinton2015distilling} to compress deep models. 
While such methods are efficient in terms of compression rate, deep model compression necessarily trade-offs the accuracy.
Rather than compressing the model with accuracy losses, we could construct an algorithm specialized for the model transmission
scenario.
Progressive transmission is a compromise policy that can be applied after the deep model compression to improve user experience.

\subsection{Deep Model over a Network}

Recently, Chen \textit{et al.} \cite{chen2019CommunicationParadigm}
introduces a design paradigm for compressing and transmitting models in a network environment, though it is dedicated to the infrastructure in cloud and edge computing and they do not consider the response time or user experience.
Collaborative intelligence \cite{eshratifar2019jointdnn,eshratifar2019bottlenet,kang2017neurosurgeon} is an another paradigm in deep model deployment, which is proposed to solve the transmission time, computation load, and storage limitations of edge devices.
In collaborative intelligence, only few layers of the model are executed on the edge device and then the output from the front layers is transmitted to the server instead of heavy raw input such as images or videos \cite{eshratifar2019jointdnn}.
Collaborative intelligence works as a compromise solution between the edge-only and the server-only approach, and the studies show that they could improve the inference latency.
However, it still requires the server resources, thus it cannot be used in the scenarios where the cost problem or the data privacy problem exists.

Unlike these methods, our progressive transmission enhances user experiences by providing approximated inference results during the transmission. In addition, our method can deal with the server cost problems or data privacy issues since any computations on the server-side are not required.

\section{Framework Design}
\label{section:framework_design}

\begin{figure*}[t]
    \centering
    \includegraphics[width=0.75\textwidth]{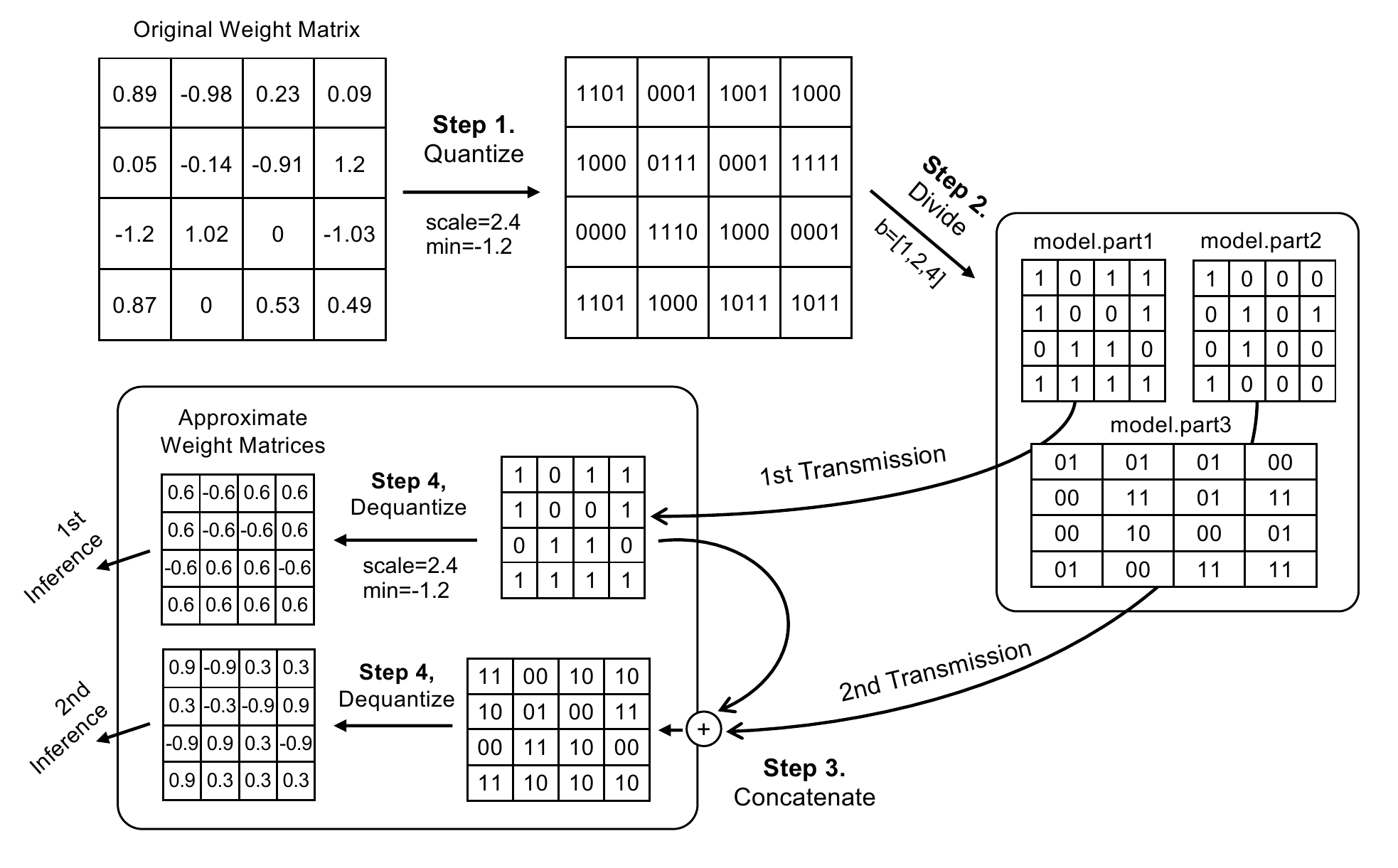}
    \caption{Overview of our framework for progressive transmission using quantization.}
    \label{fig:framework}
\end{figure*}

In this section, we propose a framework which transforms a trained model to the progressive model and provides approximate inference results during transmission.
The framework includes a representation of the progressive model, which supports multiple inferences during transmission while preserving the model size.
The overall flow of the framework is illustrated in Fig. \ref{fig:overview}.
To increase usability of progressive transmission, we design the framework to support various vision tasks, including image classification and object detection.
We further expose flexible configuration on our framework to allow the users of the framework to set the numbers of divisions and the size of each part on their demands.

\subsection{A Naive Approach}

A naive idea to design a representation for progressive transmission is to use the precision of the floating points in the model.
Because most deep learning models operate through matrix additions and multiplications, model inference works well even if the floating points are manipulated and precision is slightly decreased \cite{Gong2014Quantization}.
Therefore, by transmitting significant bits of the floating points first and transmitting trivial bits later, the intermediate results can be inferred even if only a part of the model is delivered.
For example, when the model is transmitted in two stages, a number in the model can be represented in the following form:
\begin{equation}
    1.2345678 = \underbrace{1234}_\text{1st significand} \times 10\!\!^{\overbrace{-3}^\text{exponent}} +
    \underbrace{0.5678}_\text{2nd significand} \times 10\!\!^{\overbrace{-3}^\text{exponent}}
\end{equation}
where the first significand and exponent are sent first and the second significand is sent later.
This methodology is intuitive, however, it is not efficient in terms of representation space.
Instead of splitting the floating points by digit, compression algorithms can be used to represent the model with lower precision drops.
Among them, quantization \cite{Gong2014Quantization,han2016deepcomp,Jacob2018QuantizationForIAI} is a general way to compress the deep learning models.

\subsection{Using Quantization}

Fig. \ref{fig:framework} illustrates an overview of the procedures from the original weight matrix to the intermediate inference results.
In overall, the procedure consists of four steps as follows: 1) quantization, 2) bit division, 3) bit concatenation, and 4) dequantization.
We describe the framework step-by-step in this section.

\para{Step 1: Quantization}

To better represent the deep model, we first quantize every floating point of matrices in the deep model.
The most common quantization algorithm in deep model is to calculate the maximum and minimum values for each matrix and divide the range into uniform intervals to map the matrix values.
Our quantization scheme is similar to the algorithm, which is also used in Tensorflow Lite \cite{TensorFlowLite} and Tensorflow JS \cite{TensorFlowJS} frameworks, yet we replace the rounding function with the flooring function.
Jin \textit{et al.} \cite{jin2020adabits} already have addressed the issue that the rounding function causes precision losses in bit concatenation and requires extra storage to avoid the losses.
We quantize the floating-point scalar $M_{ij}$ in matrix $M$ to the $k$-bit unsigned integer using

\begin{equation} \label{eq:quantize}
   q \langle k \rangle = Q_{ij} \langle k \rangle = floor \left( 2^k \left( \frac{M_{ij} - \min{M}}{\max{M} - \min{M} + \epsilon} \right) \right)
\end{equation}
We use a small enough value $\epsilon$ to make the range of the scalars to $[0, 2^k)$ before applying the flooring function.
Using the equation, we have a quantized matrix from the original floating-point matrix as displayed in the Fig. \ref{fig:framework}.

\para{Step 2: Bit division}

To support progressive transmission, we second divide the quantized matrix to multiple fraction matrices which have a same dimension but different bit-widths.
As one of the design goals of the framework is to provide flexible configuration to users, we expose the variable $b$ which is responding to the bit-widths of the each element of divided matrices. (e.g., $b = [1, 2, 4]$ in the figure.)
After dividing the matrcies, we progressively transmit the divided model to the user divice.
The scheme to fetch the $m$-part from the $k$-bit quantized integer $q \langle k \rangle$ is given by:
\begin{equation} \label{eq:slice}
  p \langle k,m \rangle = (q \langle k \rangle \ll b_{m-1}) \ggg (k - b_m + b_{m-1})
\end{equation}
where $b_i$ is the $i$-th bit-width and $b_0 = 0$.
Note that $\ll$ and $\ggg$ are the unsigned bit-shift operations to the left and right respectively, adopted for fast calculation.

\begin{table*}[b]
\centering
\caption{
    \label{table:time_analysis} Comparison of the total execution time of progressive and singleton models.
}
\def\arraystretch{1.1}
\begin{tabular}{l r r r r r }
    \hline
    \multirow{2}{*}{\textbf{Model}}
    &\multicolumn{1}{c}{ \multirow{2}{*}{ \textbf{Size} } }
    &\multicolumn{1}{c}{ \multirow{2}{*}{ \textbf{H/W} } }
    &\multirow{2}{*}{\textbf{Singleton}}
    &\multicolumn{2}{c}{\textbf{Progressive}}
    \\
    &&&
    & \multicolumn{1}{c}{\textbf{w/o concurrent}}
    & \multicolumn{1}{c}{\textbf{w/ concurrent}}
    \\
    \hline
    MobileNetV2 \cite{sandler2018mobilenetsv2}
    & 7.1 MB & CPU & 8s & 13s (+63\%) & 8s (+0\%) \\
    MobileNetV1 \cite{Howard2017MobileNets}
    & 8.5 MB & CPU & 10s & 18s (+80\%) & 10s (+0\%) \\
    InceptionV1 \cite{Szegedy2015Inception}
    & 13.4 MB & GPU & 14s & 17s (+21\%) & 14s (+0\%) \\
    ResNet50 \cite{He2016ResNet}
    & 51.2 MB & GPU & 52s & 63s (+21\%) & 53s (+2\%) \\
    \hline
    SSDLite-MobileNetV2 \cite{sandler2018mobilenetsv2}
    & 9.3 MB & GPU & 10s & 13s (+30\%) & 10s (+0\%) \\
    SSD-MobileNetV2 \cite{sandler2018mobilenetsv2}
    & 33.8 MB & GPU & 35s & 50s (+42\%) & 35s (+0\%) \\
    \hline
\end{tabular}
\end{table*}

\para{Step 3: Bit concatenation}

The divided matrices from the previous step are deployed to the server and transmitted to the user device when the deep model is requested.
To properly infer the deep model from the divided pieces, the divided matrices should be concatenated before the inference.
For example, the second approximate model is available after concatenating the elements of the \textit{model.part1} and \textit{model.part2} matrices, grouped by local positions and restoring the floating points.
The scheme to concatenate the divided binaries is given by
\begin{equation} \label{eq:concatenate}
\begin{aligned}
  q' \langle k \rangle = (p \langle k,1 \rangle \ll (k - b_1))
  \mathbin{|} (p \langle k,2 \rangle \ll (k - b_2)) \\
  \mathbin{|} \cdots \mathbin{|} (p \langle k,n \rangle \ll (k - b_n))
\end{aligned}
\end{equation}
where the operator $|$ indicates the bitwise \textit{or} operation, and $n$ is the available number of fractions.

\para{Step 4: Dequantization}

After concatenation, a quantized integer in the matrix is restored to a floating point.
This operation is the inverse of quantization, and the equation is given by
\begin{equation} \label{eq:dequantize}
    M'_{ij} \langle k \rangle = (\max{M} - \min{M}) \cdot \frac{q' \langle k \rangle }{2^{k}} + \min{M} + \frac{1}{2^{k+1}}
\end{equation}
where $1/(2^{k+1})$ is a revised factor that restores the loss from the flooring function.

\subsection{Concurrent Transmission and Inference}
\label{section:concurrent}

While quantization and bit division are performed once before the model deployement, concatenation and dequantization are performed at every inference on the user device.
Furthermore, our framework requires multiple inferences to show the intermediate results of the model.
However, those overheads from concatenation and multiple inferences become a problem since user devices usually do not have powerful hardware, unlike the enterprise servers.
Towards minimizing the overhead from progressive inference, we introduce concurrent execution of transmission and inference.

\begin{figure}[t]
    \centering
    \includegraphics[width=0.48\textwidth]{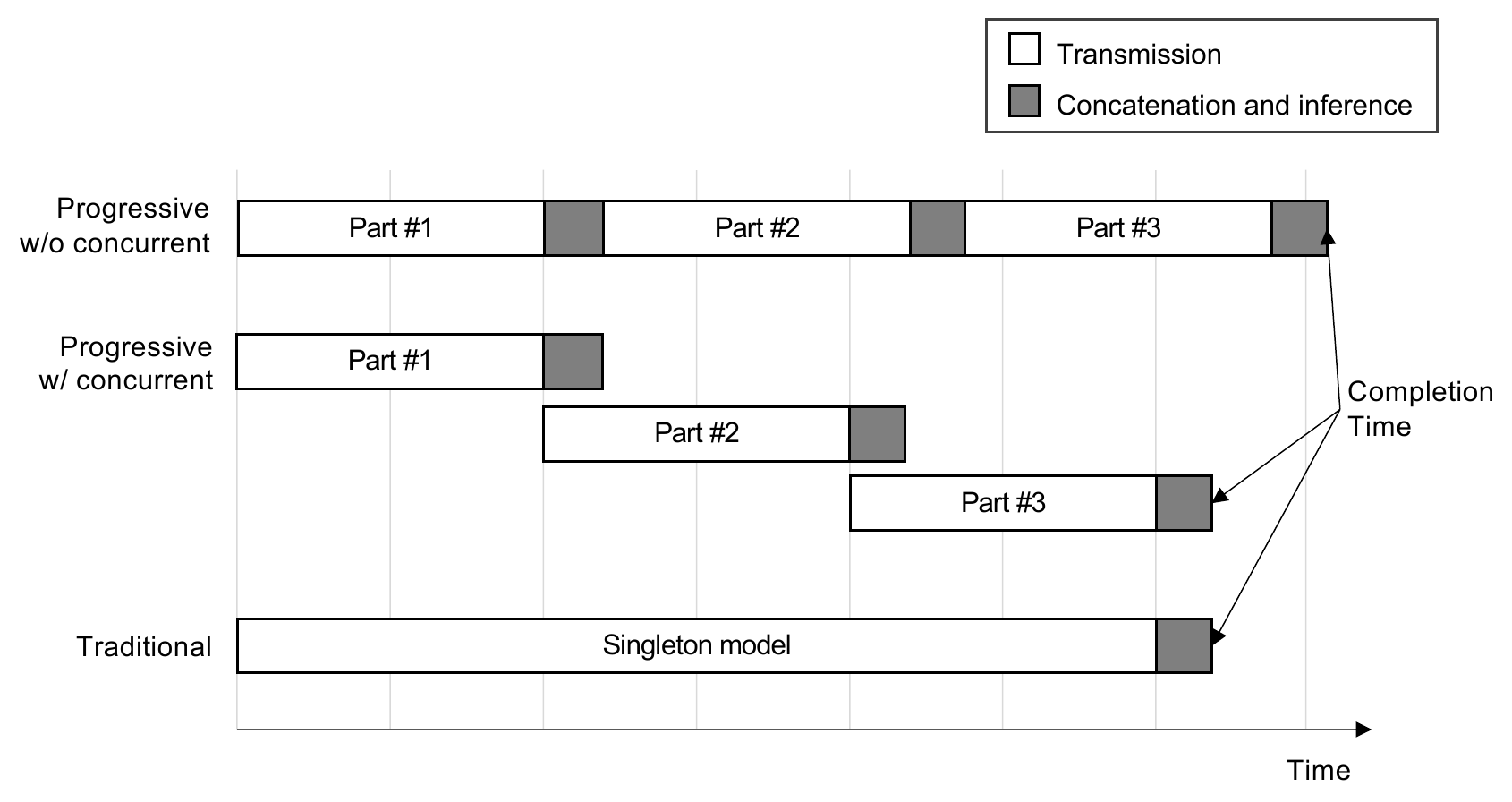}
    \caption{Timeline of singleton model transmission and progressive transmission with/without concurrent inference.}
    \label{fig:concurrent_timeline}
\end{figure}

Fig. \ref{fig:concurrent_timeline} shows timelines of a singleton model and two progressive models, where the upper model is naively implemented while the lower model performs concatenation and inference concurrently with the transmission.
When the inference is performed concurrently with the transmission, we could achieve equivalent completion time compared to the traditional method even if additional operations for concatenation and multiple inferences are added.
As modern operating systems and browsers download the files in the background, concurrent execution can be implemented easily and efficiently.

\section{Evaluation}
\label{section:evaluation}

\begin{figure*}[t]
    \centering
    \includegraphics[width=0.83\textwidth]{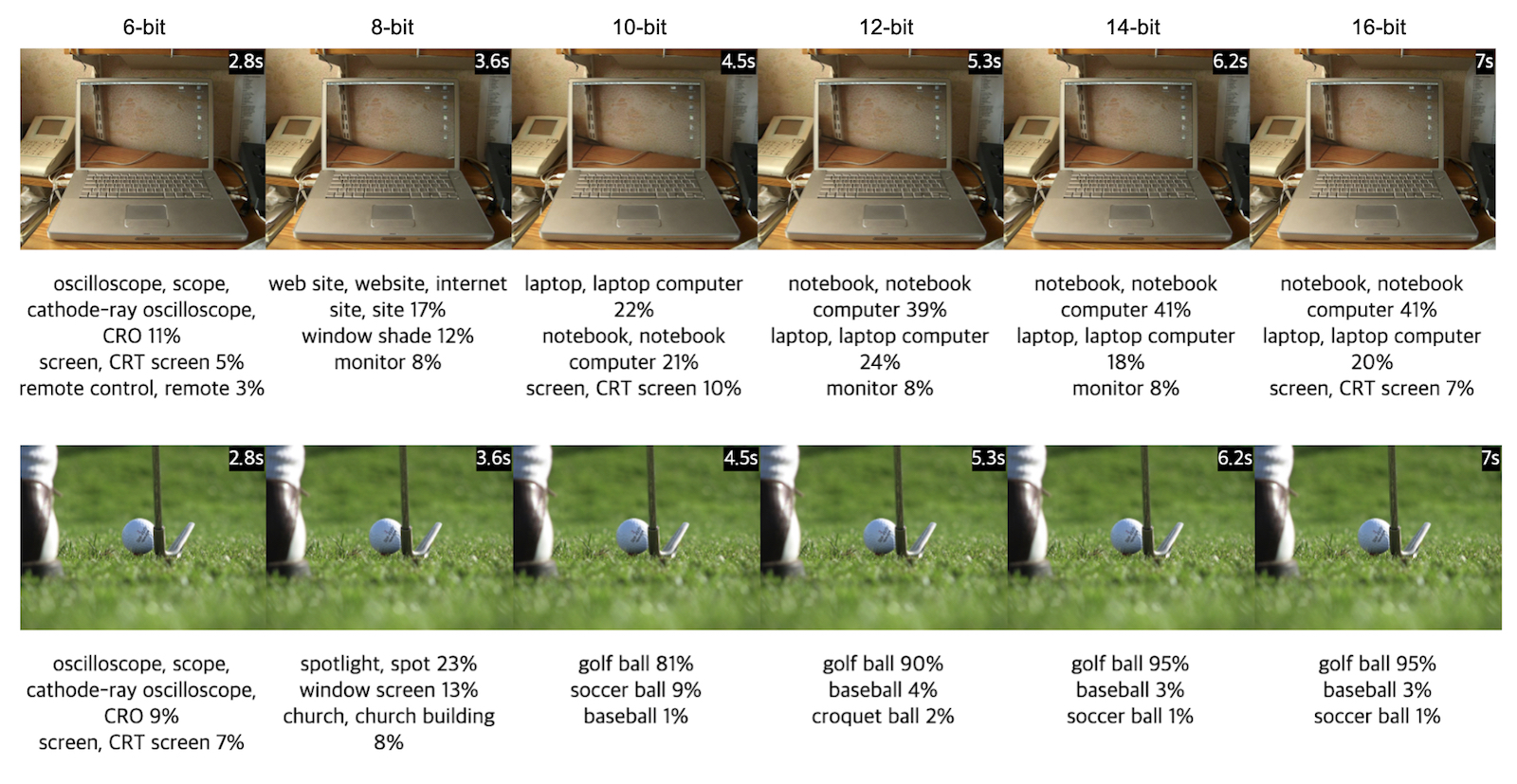}
    \caption{Intermediate results from the progressive image classification model (MobileNetV2, Transmission speed: 1.0 MB/s).
    Intermediate inference results from 2-bit and 4-bit models are omitted since the accuracy is too low at 4-bit or less bit-widths.
    }
    \label{fig:classification_example}
\end{figure*}

\begin{figure*}[t]
    \centering
    \includegraphics[width=.95\textwidth]{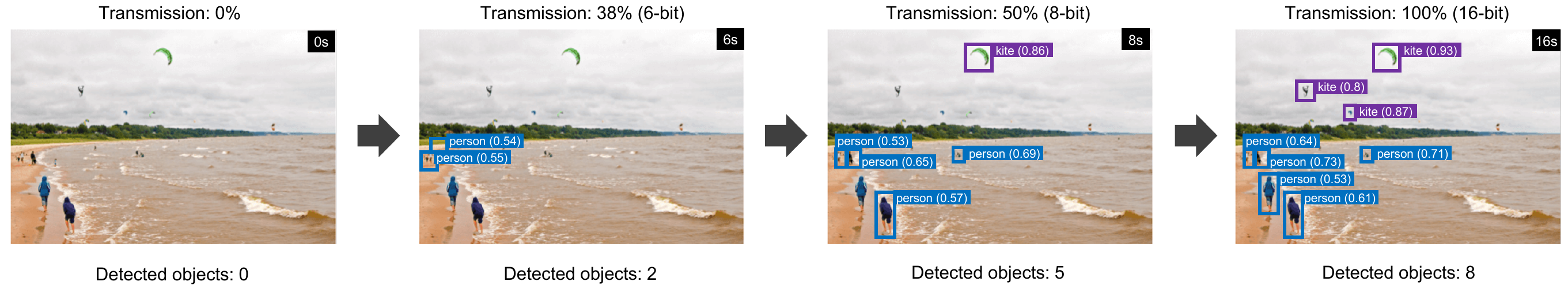}
    \caption{Intermediate results from the progressive object detection model (SSD-MobileNetV2, Transmission speed: 2.5 MB/s).}
    \label{fig:detection_example}
\end{figure*}

\subsection{Total Execution Time}

To demonstrate that our framework does not increase the total execution time, we measure the total execution times of popular deep learning models with our JavaScript implementation.
Table \ref{table:time_analysis} shows the summarized results of the concurrent transmission and inference.
The first four models are object classifiers trained with ImageNet \cite{imagenet}, and the two models below are the object detection models trained with MS COCO \cite{lin2014mscoco}.
Notably, the total execution times (transmission + concatenation + dequantization + inference) of the progressive models are equivalent to the singleton models if the concurrent execution is implemented, while an additional 20\% to 80\% longer time occurs when the concurrent execution is not implemented.

% This paper aims to improve the response time and user experience when transmitting deep models over a network.
% However, even if we provide intermediate results in the middle of the transmission, user experience might be harmed if the total execution time increases.
% To prevent the overall execution time from increasing, we introduced concurrent execution of transmission and inference and revealed that it works effectively.

\para{Experimental Setup}
With the observation that 16-bit quantized models show equivalent accuracy to the full-precision model, we used 16-bit quantized models as baselines.
We configured the transmission to proceed eight times in the progressive method (2$\rightarrow$4$\rightarrow$6$\rightarrow\cdots\rightarrow$14$\rightarrow$16), and used CPU(JS) on inferring small models and GPU(WebGL) on inferring relatively heavy models.
The experiment was conducted on Macbook Air (M1, 2020), Chrome 88, and TensorFlowJS.
Transmission speed was set to 1 MB/s.

\begin{table*}[b]
\centering
\caption{
    \label{table:accuracy} Comparison of the accuracy (\%) of progressive and singleton (orig.) models: top1 metric for object classification with three models (rows 2-4) and boxAP metric for object detection with three models (rows 5-7).
}
\def\arraystretch{1.1}
\begin{tabular}{lrrrrrrrrr} \hline
\multirow{2}{*}{\textbf{Model}} & \multicolumn{9}{c}{\textbf{Bit-width}} \\
& \multicolumn{1}{c}{\textbf{2}} & \multicolumn{1}{c}{\textbf{4}} & \multicolumn{1}{c}{\textbf{6}} & \multicolumn{1}{c}{\textbf{8}} & \multicolumn{1}{c}{\textbf{10}} & \multicolumn{1}{c}{\textbf{12}} & \multicolumn{1}{c}{\textbf{14}} & \multicolumn{1}{c}{\textbf{16}} & \multicolumn{1}{c}{\textbf{orig.}} \\ \hline
MobileNetV2 \cite{sandler2018mobilenetsv2} & 0.0 & 0.0 & 40.1 & 70.7 & 71.8 & 71.9 & 71.9 & 71.9 & 71.9 \\
ResNet18 \cite{He2016ResNet} & 0.0 & 0.0 & 67.5 & 69.6 & 69.8 & 69.8 & 69.8 & 69.8 & 69.8 \\
EfficientNet-b0 \cite{tan2019efficientnet} & 0.0 & 0.0 & 0.0 & 48.7 & 70.9 & 76.4 & 77.5 & 77.5 & 77.6 \\
\hline
SSD300-VGG16 \cite{liu2016ssd} & 0.0 & 0.0 & 18.6 & 25.0 & 25.1 & 25.1 & 25.1 & 25.1 & 25.1 \\
SSDLite320-MobileNetV3-Large \cite{howard2019mobilenetv3} & 0.0 & 0.0 & 15.5 & 20.3 & 21.1 & 21.2 & 21.3 & 21.3 & 21.3 \\
FasterRCNN-MobileNetV3-Large-FPN \cite{ren2015fasterrcnn,howard2019mobilenetv3} & 0.0 & 0.0 & 24.3 & 31.8 & 23.6 & 32.8 & 32.8 & 32.8 & 32.8 \\
\hline 
\end{tabular}
\end{table*}

\subsection{Qualitative Results}

We qualitatively show examples of progressive transmission for the classification and detection models in Figs. \ref{fig:classification_example} and \ref{fig:detection_example}, respectively.
Our progressive model provides approximate results before the entire model file is transmitted, allowing users to interact with the model quickly and improving the user experience.

\subsection{Accuracy Analysis}

% 이쪽 문단 완전히 새로 작성했는데 괜찮은지 봐주세요!

We evaluate the accuracy of progressive models with popular classification and detection model architectures.
We borrowed pre-trained models provided in online websites and converted the models with our framework to support progressive transmission.
Table \ref{table:accuracy} shows the accuracy comparison results, where we measure the ImageNet top-1 accuracy and COCO boxAP of approximate models provided by our framework during transmission.

% progressive mode를 지원하면서 size도 같은데 정확도 손실도 없다 (tradeoff가 없다는 의미)
% In final model accuracy, we reveal that there is no accuracy degradation on our progressive models even though our model has the same size compared to the original pre-trained model.
In final model accuracy, we reveal that there is no accuracy degradation on our progressive models even though our model supports progressive inference during transmission.
In intermediate model accuracy, we could not receive meaningful inference results in 2-bit and 4-bit models due to the precision loss, yet it has been shown that better inference results can be obtained from 6-bit and later.
One interesting part is that our experiments were conducted only by converting the existing pre-trained models without adaptive quantization-aware training \cite{jin2020adabits,yu2019any}, which shows high scalability of our method.

% One interesting part is that our experiments were conducted only by converting the existing pre-trained models without additional training. If we train the models to be optimized for the lower bit-widths using adaptive quantization-aware training \cite{jin2020adabits,yu2019any}, we could achieve better accuracy in lower bit-widths.

%which shows high scalability of our method. % quantization-aware training과 scalability간의 trade-off로 보고 이 기술의 high scalability를 부각하고싶으나, 인용논문의 한계를 잘 알지 못해 고민중

% One interesting part is that our experiments were conducted only by converting the existing pre-trained models without quantization-aware training.
%If we train the models to be optimized for the lower bit-widths using adaptive quantization-aware training \cite{jin2020adabits,yu2019any}, we could achieve better accuracy in lower bit-widths.

\subsection{User Study}

We further demonstrate the applicability of the progressive transmission in a real-world environment by conducting a user experiment with our web application.
The experiment compares the user's tolerance between conducting the progressive transmission or not, when the long transmission time of deep models is given.

\para{Experimental Setup}

\begin{figure}[t]
    \centering
    \includegraphics[width=.5\textwidth]{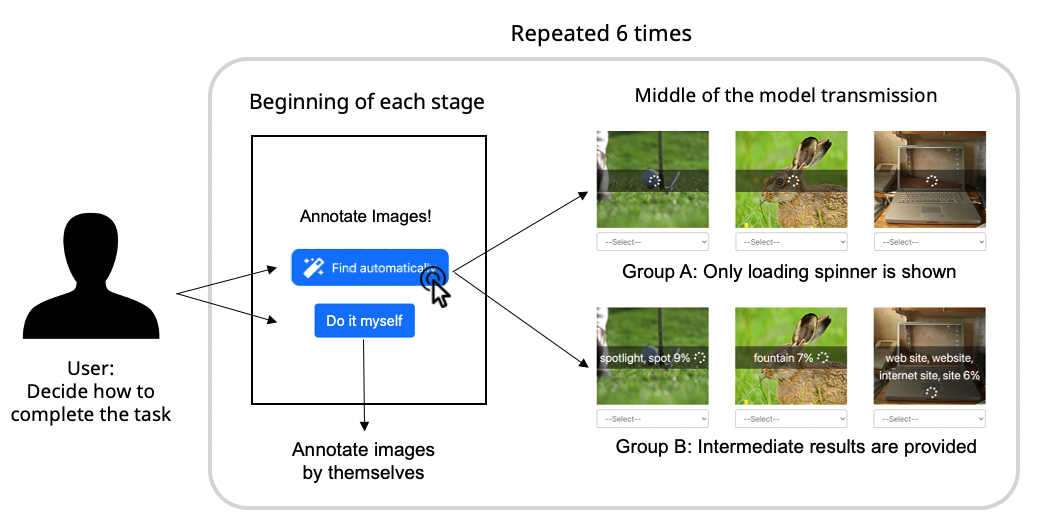}
    \caption{\label{fig:exp-concept} Overall design of the user experiment.}
\end{figure}

Fig. \ref{fig:exp-concept} shows the overall design of our user study.
First, the participants are given two choices to classify an image: obtain the classification results from a deep learning model (\textit{Find automatically} button) or manually classify without waiting for the deep model's results (\textit{Do it myself} button).
Then, we divide the participants into two groups and implement two versions of the experimental application according to the groups:
\begin{itemize}
    \item{\textbf{Group A}: No progressive transmission. Users can only see the final result after the model is fully transmitted.}
    \item{\textbf{Group B}: Progressive transmission is implemented. Users are allowed to see intermediate results before entire model is delivered.}
\end{itemize}

% In the experiments, there was no significant difference in network speed and hardware performance between Groups A and B.
We measure the ratio of the participants who actively use the \textit{Find automatically} button during the experiments.
To clearly see the effect of the progressive transmission, we limit the transmission speed of the deep model.
Due to the repetitive and boring task, the participants might prefer to use the \textit{Find automatically} button to obtain the classification results of the deep model. 
However, due to the long transmission time of the model, they may prefer to manually annotate the images instead of waiting for the model's results.
We hypothesize that Group B will be more likely to use the button than Group A, since the progressive transmission reduces the boredom and frustration caused by the slow transmission time.

\para{Implementation Detail}

The images to be classified are from the ImageNet~\cite{imagenet} validation set, and the experiment has 6 repetitive stages.
We use MobileNetV2 \cite{Sandler2018MobileNetV2} as the deep model for the \textit{Find automatically} button.
The model is executed on the web browsers of the participants' own devices (desktops or laptops).
We implement the experimental application as a web application, using TensorFlowJS \cite{TensorFlowJS}.
The model transmission speed is set to multiple configurations (0.1 MB/s, 0.2 MB/s, and 0.5 MB/s) to simulate from slow network speed to relatively fast network speed.
Participants are asked to classify 8 or 12 images in each stage (12 images on 0.1$\sim$0.2 MB/s and 8 images on 0.5 MB/s).
The network speed is evenly configured among Groups A and B.
In Group B, the model is quantized after training and then eight intermediate results are provided during transmission {\small(2$\rightarrow$4$\rightarrow$6$\rightarrow\cdots\rightarrow$14$\rightarrow$16)}.

\para{Result}

We recruited 66 participants online, and randomly distributed to Groups A and B.
A total of 57 valid data were collected, where we excluded data from the participants who have never used the \textit{Find automatically} button during the experiment.
Then we measured the ratio of participants who used the \textit{Find automatically} button greater or equal to three times during six stages (click ratio $\geq50\%$).

\begin{table}[t]
\centering
\caption{
    \label{table:ux_result1} Ratio comparison of the participants who actively used the deep learning-based automatic tool among Groups A(w/o progressive trans.) and B (w/ progressive trans.).
}
\def\arraystretch{1.1}
\setlength{\tabcolsep}{5pt}
\begin{tabular}{l | c c}
    \hline
    \multirow{2}{*}{
        \textbf{Network Speed}
    }
    & \textbf{Group A}
    & \textbf{Group B}
    \\ & ($n$=29) & ($n$=28) \\
    \hline
    0.1MB/s ($n$=18) & 44\% & 67\% \\
    0.2MB/s ($n$=23) & 42\% & 64\% \\
    0.5MB/s ($n$=16) & 50\% & 88\% \\
    \hline
    \textbf{Overall} & \textbf{45\%} & \textbf{71\%} \\
    \hline
\end{tabular}
\end{table}

Table \ref{table:ux_result1} shows the results of our experiment.
In Group A, more than half of the participants gave up on using the automatic tool and performed the task manually.
In contrast, more than 70\% of participants in Group B actively used the tool, which is a consistent result with our hypothesis.
Notably, progressive transmission was effective in all network speed configurations, showing possibility that progressive transmission might work as a general solution when delivering a model through a network.

\begin{figure}[h]
    \centering
    \includegraphics[width=.37\textwidth]{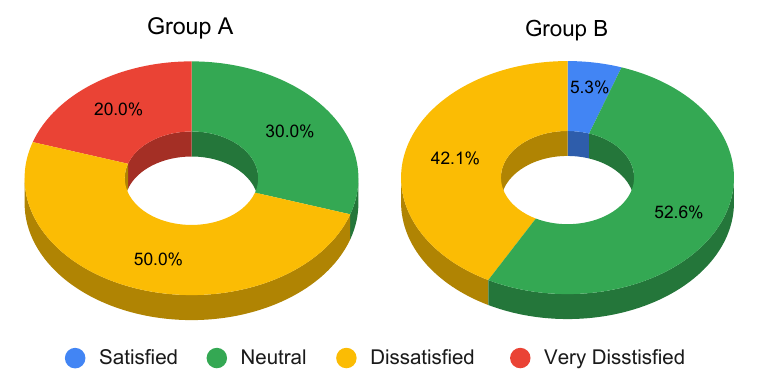}
    \caption{\label{fig:survey_pie} Survey results of the deep learning model speed after the experiment.}
\end{figure}

We further asked participants about the experience of the \textit{Find automatically} button after the experiment.
Fig. \ref{fig:survey_pie} shows the survey results, where 39 participants submitted the survey results out of 57 participants.
As a result, Group A tended to be more dissatisfied with the model's inference speed than Group B.
% Under the same conditions, participants in Group B were allowed to receive the first inference result more quickly shown in Fig. \ref{fig:response_time} (4 times faster than Group A in average). We derived that the faster response time reduced some of the dissatisfaction caused by the slow transmission speed and improved the user experience of the deep model.
We suppose that progressive transmission reduces some of the dissatisfaction caused by the slow transmission speed.

Taking the above results together, we argue that progressive transmission and inference of deep models is beneficial to the user experience.
As the size of the deep model increases or the network speed is limited, the more effectiveness of our progressive transmission will be shown.
In real-world scenario, progressive transmission could compromise the long transmission time of the deep models, especially for the countries that have not yet constructed fast networks.

\section{Conclusion}
\label{section:conclusion}

We introduced the progressive transmission and inference of deep learning models which allows approximate inference results in the middle of the transmission.
Our progressive transmission framework allows multiple inferences during the transmission without increasing the model size and total execution time.
We demonstrate that our method improves the user experiences by conducting a user study with a real world-like application.
To the best of our knowledge, this is the first approach to introduce progressive transmission into deep learning models.
We implemented progressive model transmission in a simple way in this paper, but we believe that it can be implemented with more advanced methods.
We hope that this study motivates academy and industry to consider user experience in addition to deep learning model size or the transmission time, and activates the applications and services that transmits deep learning models to user devices.

\section*{Acknowledgement}

This work was supported by Institute of Information \& communications Technology Planning \& Evaluation (IITP) grant funded by the Korea government(MSIT) (No.2019-0-01158, Development of a Framework for 3D Geometric Model Processing)

\bibliographystyle{IEEEtran}
\bibliography{ref.bib}

\end{document}